\documentclass{article}

\usepackage[preprint]{neurips_2026}

\usepackage[utf8]{inputenc} %
\usepackage[T1]{fontenc}    %
\usepackage{hyperref}       %
\usepackage{url}            %
\usepackage{booktabs}       %
\usepackage{amsfonts}       %
\usepackage{nicefrac}       %
\usepackage{microtype}      %
\usepackage{xcolor}         %
\usepackage{amsmath}
\usepackage{graphicx}
\usepackage{algorithm}
\usepackage{algpseudocode}
\usepackage{enumitem}
\usepackage{placeins}

\newtheorem{lemma}{Lemma}
\newtheorem{theorem}{Theorem}

\newcommand{\eat}[1]{}

\title{What Was That Again? Certified Robustness for Automatic Speech Recognition}

\author{%
  Andrew C. Cullen \\
  University of Melbourne \\
  \texttt{andrew.cullen@unimelb.edu.au} \\
  \And
  Neil G. Marchant \\
  University of Melbourne \\
  \And
  Jiani Xie \\
  University of Melbourne 
  \AND 
   Paul Montague \\
   DST Group, Adelaide \\
   \And
  Benjamin I. P. Rubinstein \\
  University of Melbourne \\
}

\begin{document}

\maketitle

\begin{abstract}
Automatic Speech Recognition systems are notoriously both sensitive to adversarial and benign perturbations. While this has been repeatedly demonstrated using reference datasets, detecting such behaviors in deployed systems is incredibly challenging, due to the absence of oracle knowledge of the true transcription. We demonstrate that employing a certification-inspired mechanism can significantly decrease WER, increase recall, and decrease the Spearman correlation between confidence and WER. We achieve this through a dual-gate diagnostic pipeline: a Two-Sided Atomic Audit that accumulates statistical wealth to certify both token existence and adversarial exclusion, and a Rank-Based Tournament that selects the winning sequence. Our evaluations across four diverse architectures demonstrate up to a $55\%$ relative reduction in Word Error Rate, while also providing granular word- and sentence-level certifications to enhance acoustic security.

\end{abstract}

\section{Introduction}

For all their transformative utility, neural networks in speech processing remain notoriously sensitive to infinitesimal input perturbations. These perturbations---known as \emph{adversarial examples}---demonstrate that model decision boundaries often lack the semantic alignment required for safety-critical deployments. While many approaches have been proposed to mitigate the risk associated with these adversarial examples, the most conceptually promising solution is a family of guarantees producing \emph{Certified Robustness}~\citep{lecuyer2019certified, cohen2019certified, cullen2022double}. 

These robustness mechanisms are rigorous frameworks for mathematically guaranteeing that a model's prediction remains invariant to perturbations. In the case of a classifier $F$, the guarantee is oftentimes a radius $r$ such that we can guarantee $F(x) = F(x')$ for all $x'$ in the $\ell_p$-ball $B_{p}(x,r) = \{x' : \| x' - x\|_p \le r \}$. While several approaches have been proposed, Randomised Smoothing (RS) has proven particularly effective, as it introduces no additional architectural or infrastructure burdens on the model $F$~\citep{cohen2019certified}. %

However, extending the guarantees of CR to non-categorical, high-dimensional sequence outputs---such as those encountered in Natural Language Processing or Automatic Speech Recognition (ASR)---presents a combinatorial challenge~\citep{huang2023rs, huang2024cert}. In ASR, if sentences are treated as discrete classes, then the output space explodes in size as noise levels increase. Traditional RS workflows, which rely upon finding a sentence representing the \emph{majority class} fail, because the probability mass of any single transcription collapses under such conditions~\citep{olivier2023assessing}. Resolving this limitation as crucial, as it is incredibly difficult to audit deployed acoustic models to ensure that they are producing safe outputs when exposed to untrusted inputs. 

This failure mode reveals a fundamental dual-layer opportunity for sequence certification, in that one must not only certify the \emph{atomic content}, the specific words present in the signal, but also their \emph{structural arrangement}, representing the relative ordering and grammatical coherence of these atomic elements. Prior works in this space have attempted to address this through numerically expensive sequence alignment; however the resulting outputs have typically focused upon either providing high-confidence word inclusions at the cost of structural fragmentation, or attempting holistic sequence certification that is hamstrung by the problem space's inherent combinatorial scaling. 

In this work, we propose a novel framework to address both forms of certification through E-value tournaments to produce \textbf{Rank-Based Sentence Certification}~\citep{shafer2019game, ramdas2023game}. Our approach bridges the gap between atomic and structural guarantees through the use of a dual-gate certification pipeline, that eschews the need for sentence alignment, resulting in a stable certification recall (\textbf{40.5\%--90.3\%}) even at high noise levels, where alternate baselines collapse to \textbf{$<1\%$}. Our new approach yields non-vacuous safety radii, and significant WER improvements on standard benchmarks, where traditional smoothing fails to certify. Our contributions are:%
\begin{enumerate}[leftmargin=*, noitemsep, topsep=0pt]
    \item \textbf{Anytime-valid Certification for Sequence-to-Sequence tasks}: We demonstrate how Ville’s Inequality and E-values can be deployed to provide valid safety radii at any point in the sampling process. This provides a framework for reducing the computational cost of certifications.%
    \item \textbf{Dual Gate and Two-Sided Adversarial Certifications}: We introduce a dual-gate pipeline that performs a two-sided audit—simultaneously proving the existence of safe tokens and the rigorous exclusion of adversarial hallucinations. This mechanism allows the system to prune the search space before the final rank-based tournament, which supplements the final certification with a measure of robustness to sentence-level changes.
    \item \textbf{Dynamic ASR Auditing}: Demonstrating how our dual-phase certifications can be employed to audit the performance of ASR models on a Part-of-Speech framework on LibriSpeech and Common Voice, even in extreme noise.
\end{enumerate}

\section{Related Work}\label{sec:related}

\paragraph{ASR} Modern Automatic Speech Recognition (ASR) systems can be categorized in terms of their architecture. The most commonly employed approaches include CTC-based architectures such as DeepSpeech~\cite{hannun2014deep_speech}; self-supervised Transformer models such as Wav2Vec 2.0~\cite{baevski2020wav2vec2}; and large-scale encoder-decoders such as Whisper~\cite{radford2022whisper}. However, in each of these, the model is susceptible to adversarial perturbations that can substantially degrade transcription accuracy~\cite{carlini2018audio_adversarial_examples, qin2019imperceptible_asr,olivier2022fooling_whisper}.

While these perturbations exist within the long, established, mature history of adversarial manipulations against computer vision tasks~\citep{goodfellow2014explaining, madry2017towards, cullen2024tu}---manipulating models to change predictions---the audio domain presents distinct challenges. Speech is a time-varying physical signal in which small waveform changes do not necessarily correspond to small changes in perceived sound, and commonly used $\ell_p$-norm distances can correlate poorly with human judgment of audibility~\cite{qin2019imperceptible_asr, schonherr2019psychoacoustic_hiding}. 

With this said, the most common metrics for assessing acoustic systems are the Signal-to-Noise Ratio (SNR) and the Word Error Rate (WER). The SNR serves as the fundamental measure of input perturbation, representing the log-ratio of signal amplitude to noise amplitude by
\begin{equation}
    SNR_{dB} = 20 \log_{10} \left( \frac{\|x\|_p}{\|\epsilon\|_p} \right)
\end{equation}
where $x$ is the clean audio signal, $\epsilon$ is the additive perturbation, and $p$ is the norm, which is typically $p \in \{2, \infty\}$. In adversarial contexts, attackers seek to minimize the perturbation magnitude (maximizing SNR) while inducing catastrophic failures in the model's output. 

Conversely, the WER serves as the standard metric for transcription quality, calculated as the normalized Levenshtein distance
\begin{equation}
    WER = \frac{S + D + I}{N}
\end{equation}
where $S, D,$ and $I$ represent the number of substitutions, deletions, and insertions required to align the hypothesis with a ground truth of length $N$. While it is ubiquitous in acoustic (and textual) contexts, it must be emphasized that the WER can often be distorted by the failure modes commonly seen within such systems. In recognition of the issues inherent to these metrics, alternative approaches have considered acoustic performance through a psychoacoustic lens~\cite{szurley2019perceptual_audio_attacks, sun2024commanderuap}, although it has been noted that it is fundamentally difficult to define consistent notions of imperceptibility~\cite{hussain2021waveguard}.

\paragraph{Certified Robustness}

While originally built upon foundations of Differential Privacy, the current state of the art in RS applies the Neyman-Pearson lemma. At their core, all RS based certifications transform a model $f$ into a smoothed counterpart $g$, subject to provable $\ell_p$ margin guarantees. As established by \citet{cohen2019certified}, for a noise level $\sigma$, the radius $r$ is a function of the probability $p_A$ of the \emph{most likely class}\footnote{The original formulation of \citet{cohen2019certified} was derived in terms of the two most likely class probabilities $p_A$ and $p_B$, however nearly all implementations reduce to the variant described in this paper.}. To achieve this, such certifications employ an independent two-phase approach, where Phase I employs an initial batch to establish the target class, while Phase II repeatedly samples under noise by way of $p_A = \mathbb{P}_{\epsilon \sim \mathcal{N}(0, \sigma^2 I)}[f(x+\epsilon) = c_A]$. This probability denotes the probability that the smoothed classifier predicts the target class $c_A$ under additive Gaussian noise. This ultimately yields a certification formulation of $r = \sigma \Phi^{-1}(p_A).$

While $p_A$ represents the theoretical expectation over the noise distribution, practical calculation of this quantity is impossible. While this can be practically estimated through Monte-Carlo sampling, the need to construct a conservative certification---to control the risk of the certification being violated---requires obtaining a high-probability lower bound $\underline{p_A}$ such that $P(p_A < \underline{p_A}) \le \alpha$, which is typically estimated through the Clopper-Pearson mechanism~\citep{clopper1934use}. 

The need to produce tight lower bounds is the primary driver of RS's temporal cost, which typically requires tens of thousands of samples.%
This is not to say that all inputs require such a volume of samples, but the frequentist approach of Clopper-Pearson introduces a \textbf{peeking problem}: if a practitioner monitors the empirical mean and stops sampling once an online estimate looks appropriate for certification, the Type-I error rate is no longer bounded by $\alpha$~\citep{johari2017peeking} as peeking introduces a multiple comparison problem. Thus setting the number of samples $n$ is an important-but-wasteful part of RS---setting it too high results in inputs far from the decision boundary requiring thousands of redundant model calls; while setting it too low it may lead to a total failure to certify. %

As an alternative to the computational limitations of frequentist frameworks, E-values provide an expectation-constrained hypothesis testing measure that is immune to the peeking problem. An E-value for the null hypothesis $H_0$ is a non-negative random variable $E$ such that $\mathbb{E}_{H_{0}}[E] \le 1$. For the purposes of controlling the Type-I error associated with certifications, $E$ can be viewed as a multiplier on wealth in a fair game. While such a context renders wealth a martingale, expected wealth cannot grow under the null hypothesis. Define the accumulated wealth $W_t = \prod_{i=1}^{t} E_i$.  Crucially, this wealth-testing tramework is inherently anytime-valid by Ville's inequality~\citep{ville1939etude, doob1940regularity}, which states that
\begin{equation}
    P\left(\exists t: W_t \ge \frac{1}{\alpha}\right) \le \alpha\enspace,
\end{equation}
thus allowing for optimal stopping with zero penalty. As long as the accumulated wealth $W_t$ surpasses $1/\alpha$, we can be confident that the probability of a Type-I error is strictly bounded by $\alpha$. 

The application of E-values to RS was pioneered by \citet{voravcek2024treatment}; however, their approach is rooted in classical hypothesis testing, rather than standard certification practices. Under the Voracek approach---as defined in their Definition 3.1---the tested null hypothesis is restricted to a pre-defined radius $r_0$, leading to evaluation of the null hypothesis
\begin{equation}
    H_0: p_A \le \Phi\left(\frac{r_0}{\sigma}\right)\enspace.
\end{equation}
While this yields a binary certification of whether a sample is robust to a radius of $r_0$, it fails to identify the \emph{maximum certified radius}. Naively extending this framework would require running an infinite number of independent hypothesis tests, for each possible $r_0 \in [0, \infty)$. As such testing is impossible, we instead will employ the Method of Mixtures to provide E-value based certifications, allowing infinite hypothesis testing in finite computational time. 

\paragraph{Certifications Beyond Classifiers}

The baseline for sequence certifications, as pioneered by \citet{olivier2021sequential}, relies on performing a task known as multiple sequence alignment to reduce the high-dimensional transcription space into a structured voting network. To achieve this, the system maintains a consensus backbone $\mathcal{B} = \{S_1, S_2, \ldots, S_L\}$, where each element $S_i$ is a dictionary that acts as a frequency counter of words observed at that relative temporal position. For each new sample drawn under noise $\epsilon$, the algorithm performs a string-to-graph alignment via a Word Transition Network to find the optimal mapping between the new tokens and the existing dictionaries. If a token cannot be aligned to an existing slot within a defined similarity threshold, the backbone is expanded, creating a new slot to represent this new element. 

However, as we will show within this work, such an approach is fragile to the very kinds of low SNR regimes that would likely be seen within an adversarial ASR context. Under these conditions, ASR models produce high variance outputs typified by small but frequent structural changes (manifesting as repeated syllables, misspellings, or substituted words), which rapidly induce failures within the sequence alignment algorithm. When the alignment matcher fails to recognize two semantically identical tokens as the same slot, it defaults to creating a new insertion slot. This creates a catastrophic feedback loop: as the number of slots $L$ increases, the Bonferroni-corrected budget $\alpha/L$ becomes increasingly stringent, making it statistically impossible to certify any individual slot. %
Under testing, 18 ground-truth words expanded to over thousands of slots after only a few thousand samples. The resulting sequence becomes an interleaved concatenation of fragments from different samples---e.g., if Sample 1 is \texttt{A B C} and Sample 2 is \texttt{D E F}, the misaligned output becomes \texttt{A D B E C F}). This structural collapse leads to rapid growth in the WER and vacuous certifications.

\section{Problem Definition}
We aim to extend certified robustness to sequence-valued predictions by constructing a \emph{certified transcription} $\hat{Y}$ of an audio signal $x$, such that $\hat{Y}$ is invariant to adversarial perturbations within an $\ell_2$ radius $R$ in the input space. Formally, let $\mathcal{X}$ denote the global universe of tokens, corresponding to the model's vocabulary, and let $\mathcal{X}^{\star} = \bigcup_{L=0}^{\infty} \mathcal{X}^L$ denote the set of all finite-length sequences over $\mathcal{X}$, where $\mathcal{X}^0 = \{\emptyset\}$. The base ASR model $f : \mathbb{R}^{d} \to \mathcal{X}^{\star}$ thus maps an input signal to a transcription sequence $y = (w_1, \ldots, w_L)$ where $w_i \in \mathcal{X}$. Following \citet{cohen2019certified}, a standard smoothed predictor $F : \mathbb{R}^d \to \mathcal{X}^{\star}$ would ideally select the most probable transcription under Gaussian noise:
\begin{equation}
    F(x) = \arg\max_{y \in \mathcal{X}^{\star}} \mathbb{P}\big( f(x + \epsilon) = y \big), \quad \epsilon \sim \mathcal{N}(0, \sigma^2 I)\enspace.
\end{equation}
However, in practice, the sequence space is so fragmented that the probability of any single transcription collapses toward zero under significant noise. As such, we define the smoothed predictor through a generalized hierarchical aggregator $G(\cdot)$
\begin{equation}
    F(x) = G\big( f(x + \epsilon)\big), \quad \epsilon \sim \mathcal{N}(0, \sigma^2 I)\enspace,
\end{equation}
the form of which we will now introduce, with the aim of constructing the pair $(\hat{Y}, R)$ such that 
\begin{equation}
    \mathbb{P}\Big( \exists \, \delta \in \mathbb{R}^d \;\; \text{s.t.} \;\; \|\delta\|_2 < R \;\; \text{and} \;\; F(x + \delta) \neq \hat{Y} \Big) \le \alpha\enspace,
\end{equation}
given some global error budget $\alpha \in (0,1)$. %
To achieve this, we decompose $G(\cdot)$ into two distinct stages with a partitioned error budget $\alpha = \alpha_{atomic} + \alpha_{tourn}$.
\begin{enumerate}[leftmargin=*, noitemsep, topsep=0pt]
    \item \textbf{Token-Level Atomic Certification}: We first identify a finite \textbf{Candidate Vocabulary} $\mathcal{V} \subset \mathcal{X}$ through an initial discovery phase, and then certify words based upon this Candidate Vocabulary. Specifically, if $w \in V$, then with probability at least $1-\alpha_{atomic}$, the true probability of $w$ appearing in a noisy transcription $p(w) > 0.5$. Conversely, if $w \notin \mathcal{V}$ and $w$ is not ambiguous, we certify $p(w) < 0.5$.
    \item \textbf{Sentence-Level Structural Certificate}: A sequential tournament is conducted among a set of candidate transcriptions $\mathcal{C} \subset \mathcal{V}^\star$, which have been filtered by the atomic gate to ensure structural and statistical validity.
\end{enumerate}
The final system radius $R = \min(R_{atomic}, R_{tourn})$ is the minimum radius required to violate either of these hierarchical guarantees.

\subsection{Atomic Certifications}\label{sec:atomic}

The atomic gate identifies and verifies constituent tokens in two phases: discovery, in which the candidate vocabulary $\mathcal{V}$ is identified; and audit, in which certifications are performed. 

\paragraph{Discovery Phase} We draw $N_1$ i.i.d. samples $Y_i = f(x + \epsilon_i)$, where $\epsilon_i \sim \mathcal{N}(0, \sigma^2 I)$ to identify the Candidate Vocabulary $\mathcal{V} \subset \mathcal{X}$ by $\mathcal{V} = \bigcup_{i=1}^{N_1} \{ w : w \in Y_i \}$.
The intent of this stage is to prune the search space to tokens with non-trivial support in the local noise distribution. These samples are subsequently discarded to ensure statistical independence. However, it must be noted that any word that is not present in the first $N_1$ samples will be unable to be verified---even if it would otherwise present as a high frequency inclusion---and as such will be filtered out from all subsequent stages.%

\paragraph{Audit Phase} Following the discovery of the candidate vocabulary $\mathcal{V}$, we verify the existence of each token using a fresh stream of $N_C$ samples. For any token $w \in \mathcal{V}$, let $p_w = \mathbb{P}(w \in f(x + \epsilon))$ denote its marginal probability of appearance. We accumulate two-sided wealth via non-negative martingales $E_{pos}(w)$ and $E_{neg}(w)$ to test the null hypotheses $H_{pos}: p_w \le 0.5$ and $H_{neg}: p_w \ge 0.5$ respectively. These hypotheses will allow us to comprehensively certify the inclusion or exclusion within the observation set. 

To achieve this, for each noisy sample $t$, let $W_{w,t} \in \{0, 1\}$ be the indicator that token $w$ is present in the transcription. To ensure numerical stability and support high-throughput GPU vectorization, we perform wealth compounding in log-space. For a betting parameter $\lambda \in (0, 2]$, the accumulated wealth at time $T$ is defined as
\begin{equation}
    \ln E_{pos, T}(w) = \sum_{t=1}^T \ln \big(1 + \lambda(W_{w,t} - 0.5)\big) \qquad ln E_{neg, T}(w) = \sum_{t=1}^T \ln \big(1 + \lambda(0.5 - W_{w,t})\big)\enspace,
\end{equation}
where $E_{i,0} = 1$ for $i \in \{pos, neg\}$. Under the null hypothesis $p_w = 0.5$, the expectation of the multiplier $E_t = 1 + \lambda(W_{w,t} - 0.5)$ should be exactly $1$, rendering $E_T$ a martingale. 

As was discussed in Section~\ref{sec:related}, by Ville's inequality~\citep{ville1939etude, doob1940regularity}, the probability that the wealth ever exceeds the safety threshold is bounded through $P(\exists T : E_T \ge 1/\alpha) \le \alpha$. We thus define the \textbf{Certified Vocabulary} $\mathcal{V}{cert}$ and \textbf{Excluded Vocabulary} $\mathcal{V}{excl}$ as the sets of tokens whose wealth crosses the significance threshold $1/\alpha_{atomic}$.

\paragraph{Confidence Sequences and Certifications} To translate these measures of statistical wealth into certified radii, we must identify the set of success probabilities $p$ that are consistent with the observed evidence. Under the framework of \textbf{Confidence Sequences} \citep{ramdas2023game}, we obtain anytime-valid bounds by inverting the E-value process. For tokens in $\mathcal{V}_{cert}$ or $\mathcal{V}_{excl}$, the anytime-valid bounds are the extrame probabilities consistent with the wealth
\begin{equation}
    \underline{p}_{w,T} = \inf \{ p \ge 0.5 : L_T(p) < \alpha_{atomic}^{-1} \}, \quad \overline{p}_{w,T} = \sup \{ p \le 0.5 : L_T(p) < \alpha_{atomic}^{-1} \}
\end{equation}
where $L_T(p) = \prod_{t=1}^T \frac{\text{Bernoulli}(W_{w,t}; p)}{\text{Bernoulli}(W_{w,t}; 0.5)}$ is the likelihood ratio martingale, and $\underline{p}_w$ and $\overline{p}_w$ are the anytime-valid lower and upper bounds respectively.
Following \citet{cohen2019certified}, we map these probability bounds to a a safety radius $r_w$ in the input $\ell_2$ space. This radius represents the minimum perturbation required to alter the inclusion or exclusion of token $w$ by way of
\begin{equation}
  r_w = 
  \begin{cases} 
    \sigma \Phi^{-1}(\underline{p}_{w,T}) & \text{for } w \in \mathcal{V}{cert} \\
    \sigma \Phi^{-1}(1 - \overline{p}_{w,T}) & \text{for } w \in \mathcal{V}{excl}\enspace.
  \end{cases}
\end{equation} 
The global radius $R_{atomic} = \min_{w \in \mathcal{V}_{cert} \cup \mathcal{V}_{excl}} r_w$ represents the smallest perturbation required to alter the inclusion or exclusion of any token in the atomic gate. This ensures that any perturbation $\delta$ with $\|\delta\|_2 < R_{atomic}$ is guaranteed, with probability $1-\alpha_{atomic}$, to leave the certified and excluded vocabularies unchanged, preserving the downstream structural tournaments candidate pool.%

\paragraph{Limitations: Multiplicity and Global Control} 
It may be notable that our process applies a fixed threshold of $1/\alpha_{atomic}$ to each token independently. This provides a rigorous \textbf{local} guarantee for each word but does not strictly control the Family-Wise Error Rate over the entire vocabulary. The true Type I error rate associated with the atomic gate is not $\alpha_{atomic}$, but rather it scales with the size of the vocabulary, to having a true confidence of  $|\mathcal{V}| \alpha_{atomic}$. As such, it is possible that an adversary may be able to insert or remove a word into the certified vocabulary $\mathcal{V}_{cert}$ with a smaller perturbation than the one calculated---which would be required to survive tournament certification.%

While more conservative global control can be achieved via the e-Benjamini-Hochberg (e-BH, \citet{benjamini1995controlling, wang2022false}) procedure or Bonferroni corrections, we emphasize that the subsequent \textbf{Sentence Tournament} acts as a robust secondary gate. Even if a low-probability hallucination passes the atomic certification, it must still prove its structural and statistical validity in the tournament to appear in the final output.

For reference, the e-BH procedure would offer a more sophisticated mechanism for managing the global False Discovery Rate. By sorting the E-values $\{E_{(1)} \ge E_{(2)} \dots \ge E_{(|\mathcal{V}|)}\}$ and finding the largest $k$ such that $\frac{1}{|\mathcal{V}|} \sum_{i=1}^k E_{(i)} \ge 1/\alpha_{atomic}$, e-BH provides a rigorous guarantee that is robust to the unknown dependence structures common in ASR output distributions. We exclude e-BH from our implementation because its dynamic, data-dependent threshold is difficult to map back to stable safety radii, and the sort-and-sum operation introduces a GPU synchronization bottleneck.%

\subsection{Tournament Certification}

Up to this point, we have a certified vocabulary $\mathcal{V}_{cert}$ , which can be used to understand the components of a sentence that may be more (or less) vulnerable to adversarial manipulation. While this can provide valuable security insights in high stakes transcription environments, it still represents a bag-of-words that is unable to recreate the desired output of an ASR model: a clean, certified transcription. We bridge this gap by constructing a set of candidate transcriptions $\mathcal{C}$ restricted to the certified vocabulary $\mathcal{V}_{cert}$.

\paragraph{Nomination via Filtered Mapping} 
To achieve this, we draw a new set of $N_3$ samples under the same gaussian noise distribution as above, and apply a filtering mapping $\Pi_{\mathcal{V}_{cert}} : \mathcal{X}^\star \to \mathcal{V}_{cert}^\star$, which converts any $Y_i = (w_{i,1}, \dots, w_{i,L_i})$ to
\begin{equation}
    \tilde{Y}_i = \big( w_{i,j} \in Y_i \;\; \text{s.t.} \;\; w_{i,j} \in \mathcal{V}_{cert} \big), \quad i = 1, \dots, N_T\enspace.
\end{equation}
By removing tokens that were unable to be verified by the atomic gate, we prune hallucinations and temporal jitter. The top-$K$ most frequent unique cleaned sequences form the Structural Candidate Pool $\mathcal{C} = \{C_1, \dots, C_K\}$.

\paragraph{The Competitive Structural Tournament}

Following the filtered sampling stage, our goal is to estimate the most likely transcription under the induced distribution over filtered sequences%
\begin{equation}
    \hat{Y} = \arg\max_{y \in \mathcal{V}_{cert}^\star} \mathbb{P}\big(\Pi_{\mathcal{V}_{cert}}(f(x+\epsilon)) = y\big)\enspace.
\end{equation}
Accordingly, the tournament procedure constructs a Monte Carlo estimator $\hat{Y}$ of the smoothed predictor $F(x)$ by approximating the maximizer of the induced distribution over filtered sequences. The tournament stage maintains competing hypotheses over $\mathcal{C}$ and performs an anytime-valid sequential test. For a fresh batch of $N_T$ transcriptions $Y_t$, we update $K$ parallel E-values using a competitive betting function
\begin{equation}
  E_{i, t} = E_{i, t-1} \times \left(1 + \lambda \left(\mathbb{I}(i = \arg\min_j \text{WER}(C_j, Y_t)) - \frac{1}{K}\right)\right)\enspace,
\end{equation}
where $\text{WER}(\cdot,\cdot)$ denotes the word error rate. Under the null hypothesis that no candidate is dominant, each $E_i$ is a martingale.

\begin{lemma}[The Structural Multiplicity Subsidy]
The average wealth $\bar{E}{tourn} = \frac{1}{K} \sum{i=1}^K E_i$ is a non-negative martingale. We stop the tournament at any time $\tau$ where $\bar{E}_{\tau} \ge
1/\alpha_{tourn}$. This allows the winning candidate's wealth to subsidize the testing debt of other elements, enabling sequence-level certification with $O(1)$ multiplicity scaling relative to the candidate pool size.
\end{lemma}

The structural radius is one again produced through a \citet{cohen2019certified} style certification. However, as the tournament exists over $K$-classes (corresponding to the $K$ highest frequency observations), the associated certified radius---expressed in terms of the probability bounds $\underline{p}$ and $\overline{p}$---becomes
\begin{equation}
    R_{tour} = \frac{\sigma}{2} (\Phi^{-1}(\underline{p}_{winner}) -
      \Phi^{-1}(\overline{p}_{runner-up}))\enspace.
\end{equation}

\begin{theorem}[End-to-End Certified Transcription]
Let $\hat{Y}$ be the output of the hierarchical aggregation procedure described above, with error budget $\alpha = \alpha_{atomic} + \alpha_{tourn}$. Then, with probability at least $1 - \alpha$, the prediction $\hat{Y}$ is invariant under all perturbations $\delta$ satisfying $\|\delta\|_2 < R$, where
\begin{equation}
    R = \min(R_{atomic}, R_{tourn}).
\end{equation}
\end{theorem}

\paragraph{Discussion: Avoidance of Alignment} The traditional approach to sequence certification relies on reassembling transcriptions via sequence alignment algorithms such as Recognizer Output Voting Error Reduction (ROVER) \citep{fiscus1997post, haihua2009efficient} or confusion network lattices \citep{mangu2000finding}, and applying union bounds over the resulting paths \citep{olivier2021sequential}. However, this reassembly introduces a fundamental \emph{Multiplicity Bottleneck}. To certify a sequence of $N$ tokens, one must implicitly or explicitly prove the correctness of the underlying total order. Because a total order is uniquely defined by the set of all $\binom{N}{2}$ pairwise relations, maintaining a global confidence level $1-\alpha$ under a standard union bound requires each pairwise test to satisfy an error probability of $O(\alpha/N^2)$. This quadratic decay in statistical power rapidly results in vacuous radii. %

Moreover, alignment based mechanisms have the potential to produce transcriptions that are locally robust, but globally incoherent based upon the combination of words from multiple exclusive linguistic paths. By contrast, our framework treats the sequence as the fundamental unit of competition. By restricting the candidate pool $\mathcal{C}$ to model-generated sequences passing through the Atomic Gate, we ensure that the tournament is over linguistically coherent sentences grounded in certified evidence.%

Our categorical tournament approach transforms the sequence certification problem from a combinatorial search over an alignment lattice into a competitive wealth redistribution task. A key property of E-values is that the statistical evidence required to crown a structural winner does not scale with the number of possible word orderings, but rather with the relative dominance of the winning candidate over its primary competitors. This allows us to produce rigorous, sentence-level certificates in high-noise environments.%

\section{Analysis}

To assess the performance our approach, we considered experiments using LibriSpeech and Common Voice exposed to the Whisper-v3-Large, Whisper-Small, wav2vec 2.0 Large and HuBERT architectures (further detailed in Appendix~\ref{app:algorithm}. At a high level, the empirical results contained within Tables~\ref{tab:final_master} and \ref{tab:pos_analysis} provide strong evidence for the effectiveness of our hierarchical E-value framework to produce diagnostic markers of robustness.

\begin{figure}[t]
    \centering
    \includegraphics[width=\columnwidth]{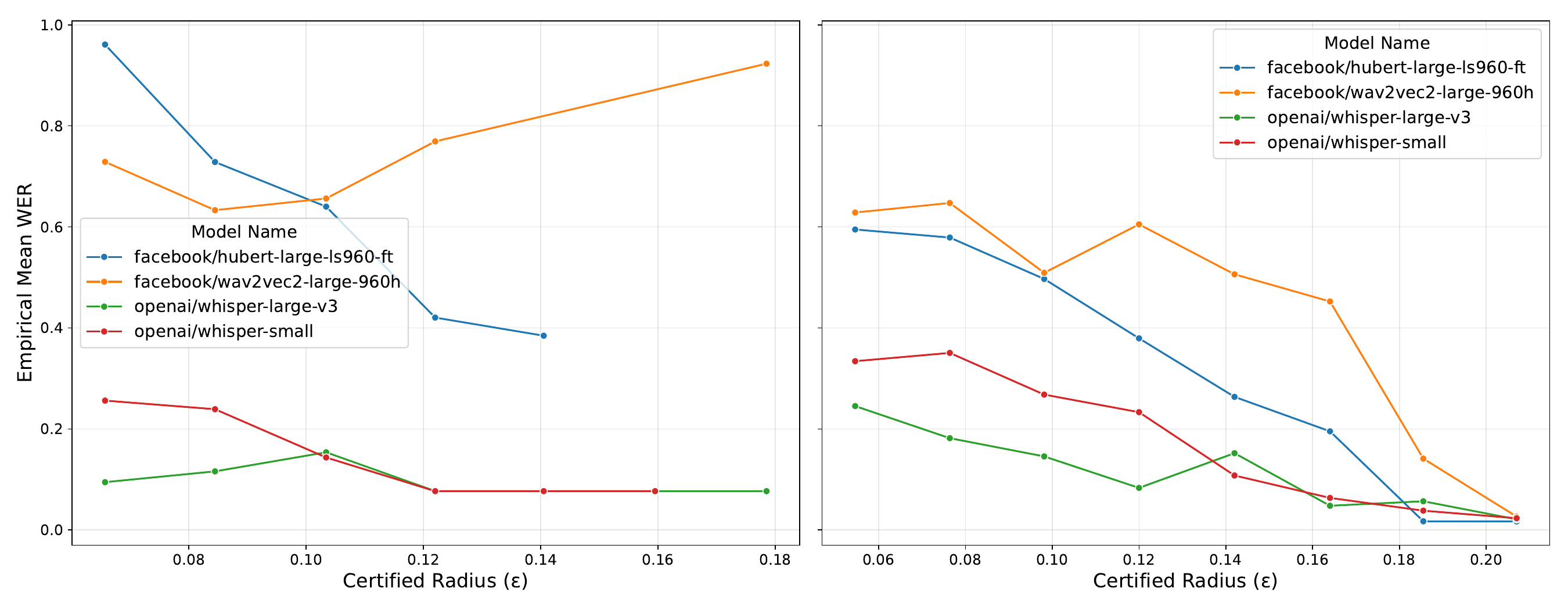}
    \caption{\textbf{Observed WER as a function of Certified Radius}: demonstrating the broad correlation between these two quantities. Left: LibriSpeech. Right: Common Voice.}
    \label{fig:correlation}
\end{figure}

Perhaps most crucially, Figure~\ref{fig:correlation} demonstrates that there is a clear correlation between the constructed Certifications and the measured WER. For the purposes of a system in production, the WER requires knowing the ground truth transcription---something that is not possible in practice. That the Certified Radius correlates with this, without requiring access to the ground truth transcription, highlights how our approach can be used to guide a confident view of the performance of an ASR system exposed to untrusted data. Figure~\ref{fig:performance} further expands on this, demonstrating that there is a notable performance advantage induced by our  mechanism. This is achieved through establishing consensus through an alignment lattice--a process that becomes statistically and computationally prohibitive as noise increases, due to the growth in hallucinated textual observations---we instead focus upon realizable computational utility through our approach. Through our atomic gate, we prune the hypothesis space before reassembly occurs, reducing sentence diversity to a managable level.%

As we noted in the Limitations section of Section~\ref{sec:atomic}, it is important to consider these certifications as a diagnostic marker, rather than a measure of true robustness, as the atomic certifications underestimate the true Type I error rate associated with the individual word certifications. However, even with this limitation, our two phase approach provides significant utility, in that it allows granular word-by-word certifications to be produced, analyzed, and used in concert with broader sentence level certifications. These have the potential to provide significant downstream utility, as it provides a mechanism for analyzing the stability of the ultimate utterance, as well as all its components. 

These results are expanded upon in Table~\ref{tab:final_master}, which demonstrates that our certified pipeline induces a significant reduction in the WER across all tested SNRs. In the highest-noise regimes, OVER (Olivier) recall collapses to \textbf{0.0--2.1\%} in extreme noise. In contrast, our framework maintains a stable certification recall of \textbf{40.5--74.0\%} at the same noise level. Furthermore, for the SOTA Whisper-Large-v3 model, our framework repairs the raw prediction from \textbf{0.273 to 0.126 WER}—a $54\%$ relative improvement in robust ASR. Notably, the observed Average Radius is inversely correlated with the SNR, validating that our anytime-valid bonds adapt to the underlying quality of the signal. That our aggregate consistently reduces the raw WER demonstrates that our approach is not merely a filter, but a robust aggregator that extracts semantic coherence from noisy model outputs.

The comparison between our Tournament framework, ROVER (Olivier), and Naive Cohen randomized smoothing reveals two critical statistical phenomena that justify the utility of anytime-valid sequence certification. As shown in Table~\ref{tab:final_master}, baseline methods (ROVER and Cohen) occasionally exhibit more negative Spearman correlation coefficients ($\rho$) than our approach at high SNR levels (e.g., ROVER's -0.825 for Whisper-Large at 10dB). However, correlation is meaningless in the presence of low recall. For example, at 10dB SNR, our framework achieves significantly higher Recall (\textbf{73\%}) than ROVER (\textbf{59\%}) while maintaining a highly informative radius ($\rho = -0.310$). Crucially, our Certified Radius provides a trust score for every sample, not just the certified subset, providing actionable information across the entire dataset. 

We also stress that as the level of induced noise increases to SNR -5.0, the utility of baseline certificates collapses entirely. That baseline recall collapses towards 0\% in noise (e.g., 0\% for HuBERT and Wav2Vec2), their certificates become a constant vector of zeros. %
In contrast, our Tournament maintains stable, non-zero variance and significant informative power even at -5dB, proving it is the only viable path for trust-scoring in extreme environments. Even relative to the raw WER, our tournament approach yields a $55.1\%$ reduction in the WER. 

\begin{table*}[t]
\centering
\caption{Comprehensive Evaluation over all SNR levels. \textit{Recall} is 99\% confidence certification success. \textit{Corr} ($\rho$) is the Spearman correlation between the method's confidence metric and the resulting WER. Note that baselines vanish in noise while our framework remains informative. Models are Hubert (H), Whisper-Large (W.L), wav2vec Large (W2.L) and Whisper-Small (W.S).}
\label{tab:final_master}
\begin{small}
\begin{tabular}{ll cccc ccc ccc}
\toprule
& & \multicolumn{4}{c}{Word Error Rate (WER $\downarrow$)} & \multicolumn{3}{c}{Recall (\% $\uparrow$)} & \multicolumn{3}{c}{Correlation ($\rho \downarrow$)} \\
\cmidrule(lr){3-6} \cmidrule(lr){7-9} \cmidrule(lr){10-12}
Model & SNR & Raw & Cohen & ROV & \textbf{Ours} & Cohen & ROV & \textbf{Ours} & Cohen & ROV & \textbf{Ours} \\
\midrule
\textbf{H} & 10.0 & 0.161 & 0.162 & 17.927 & \textbf{0.145} & 51.3 & 48.0 & \textbf{60.1} & -0.747 & \textbf{-0.845} & -0.774 \\
&  5.0 & 0.231 & 0.217 & 41.904 & \textbf{0.205} & 37.2 & 21.0 & \textbf{58.7} & \textbf{-0.756} & -0.626 & -0.731 \\
&  0.0 & 0.476 & 0.436 & 258.566 & \textbf{0.417} & 7.7 & 2.3 & \textbf{80.9} & \textbf{-0.700} & -0.258 & -0.270 \\
& -5.0 & 0.940 & 0.925 & 249.886 & \textbf{0.895} & 0.7 & 0.0 & \textbf{40.5} & -0.298 & -- & \textbf{-0.198} \\
\addlinespace
\textbf{W.L} & 10.0 & 0.085 & 0.087 & 0.970 & \textbf{0.087} & 11.0 & 44.3 & \textbf{38.2} & -0.392 & \textbf{-0.844} & -0.811 \\
&  5.0 & 0.072 & 0.052 & 1.632 & \textbf{0.051} & 8.2 & 32.5 & \textbf{90.3} & -0.348 & \textbf{-0.739} & -0.574 \\
&  0.0 & 0.081 & 0.064 & 4.317 & \textbf{0.064} & 3.5 & 15.8 & \textbf{84.0} & -0.313 & \textbf{-0.524} & -0.315 \\
& -5.0 & 0.273 & 0.125 & 52.577 & \textbf{0.126} & 0.5 & 2.1 & \textbf{74.0} & -0.154 & -0.242 & \textbf{-0.123} \\
\addlinespace
\textbf{W2} & 10.0 & 0.265 & 0.272 & 51.548 & \textbf{0.229} & 43.3 & 29.0 & \textbf{72.7} & \textbf{-0.710} & -0.708 & -0.697 \\
&  5.0 & 0.378 & 0.348 & 155.165 & \textbf{0.338} & 20.7 & 10.3 & \textbf{71.3} & \textbf{-0.752} & -0.509 & -0.476 \\
&  0.0 & 0.749 & 0.710 & 296.782 & \textbf{0.708} & 2.0 & 0.3 & \textbf{70.3} & \textbf{-0.617} & -0.099 & 0.393 \\
& -5.0 & 0.948 & 0.929 & 110.595 & \textbf{0.930} & 0.0 & 0.0 & \textbf{45.2} & -0.221 & -- & \textbf{0.008} \\
\addlinespace
\textbf{W.S} & 10.0 & 0.071 & 0.058 & 1.532 & \textbf{0.058} & 0.0 & 37.5 & \textbf{56.7} & 0.009 & \textbf{-0.711} & -0.600 \\
&  5.0 & 0.088 & 0.061 & 3.917 & \textbf{0.061} & 0.0 & 26.2 & \textbf{87.3} & -0.096 & \textbf{-0.613} & -0.608 \\
&  0.0 & 0.213 & 0.110 & 28.143 & \textbf{0.127} & 0.0 & 6.7 & \textbf{76.3} & -- & -0.372 & \textbf{-0.456} \\
& -5.0 & 0.595 & 0.669 & 258.181 & \textbf{0.443} & 0.0 & 0.0 & \textbf{44.2} & -- & -- & \textbf{-0.320} \\
\bottomrule
\end{tabular}
\end{small}
\end{table*}

\paragraph{Content Fragility}

One advantage of considering ASR transcriptions on both a sentence level and as an assembly of vocabularies is that we can consider the performance of specific textual components within the overall robustness framework. To achieve this, we employed the spaCy Natural Language Processing framework~\citep{honnibal2020spacy} to perform Part-of-Speech (POS) tagging over the transcription corpus. Table~\ref{tab:pos_analysis}'s audit reveals a crucial linguistic insight: \textbf{robustness is not uniformly distributed across word classes}. Content-heavy tokens such as nouns (NOUN) and verbs (VERB) exhibit significantly lower raw accuracy and smaller certification margins, as compared to more functional tokens (like coordinating conjunctions CCONJ and determiners DET). This is an important insight for ASR validation, that is likely explained by the relative paucity of specific nouns and proper nouns (PROPN) in the ASR training corpora.

The Certified Accuracy column demonstrates the utility of our tournament stage, in that the system can successfully recover these forms of content  words with high accuracy---including reaching $92.2\%$ accuracy for nouns, a relative improvement of $20.5\%$. This demonstrates that anchoring the structural selection to a robust functional skeleton can significantly improve the performance of ASR systems within complex acoustic environments. 

Further results can be found in Appendix~\ref{app:further_results}, covering computational efficiency and generalization.

\begin{table}
\caption{Linguistic Fragility and Accuracy by POS (see Appendix~\ref{app:POS} for details). Comparison of raw model recall vs. Certified System Recall.}
\label{tab:pos_analysis}
\begin{center}
    \begin{tabular}{llllll}
\toprule
Word Type (POS) & Raw Acc. & Cert. Acc. & Prop. Accepted & Prop. Rejected & Prop. Ambiguous \\
\midrule
CCONJ & 0.97 & 0.994 & 0.835 & 0.141 & 0.0236 \\
DET & 0.96 & 0.986 & 0.444 & 0.5 & 0.0556 \\
SCONJ & 0.919 & 0.986 & 0.74 & 0.231 & 0.029 \\
ADP & 0.932 & 0.982 & 0.747 & 0.229 & 0.0242 \\
ADV & 0.903 & 0.971 & 0.695 & 0.282 & 0.0231 \\
PRON & 0.931 & 0.967 & 0.775 & 0.201 & 0.0235 \\
VERB & 0.832 & 0.96 & 0.554 & 0.412 & 0.0341 \\
ADJ & 0.839 & 0.947 & 0.594 & 0.379 & 0.0268 \\
NOUN & 0.765 & 0.922 & 0.455 & 0.508 & 0.0365 \\
INTJ & 0.889 & 0.903 & 0.72 & 0.257 & 0.0228 \\
AUX & 0.841 & 0.899 & 0.704 & 0.268 & 0.0273 \\
PART & 0.784 & 0.795 & 0.83 & 0.148 & 0.0221 \\
PROPN & 0.539 & 0.662 & 0.296 & 0.67 & 0.0338 \\
NUM & 0.561 & 0.639 & 0.603 & 0.347 & 0.0494 \\
\bottomrule
\end{tabular}
\end{center}
\end{table}

\section{Conclusion}

In this work, we have demonstrated that acoustic robustness for sequence-to-sequence systems can be achieved through flexible, computationally efficient statistical mechanisms. By replacing combinatorial sequence alignment with a hierarchy of E-value tournaments, we demonstrate that it is possible to achieve superior performance as compared to alternate approaches, maintaining a stable Precision-Recall balance even as environmental noise increases. 

Rather than strictly optimizing for mathematically rigorous, worst-case certifications---which often scale poorly---our Tournament approach leverages these statistical mechanics to repair model transcriptions and generate dynamic safety markers. Crucially, we demonstrate that these markers strongly correlate with Word Error Rate. This correlation is invaluable for deployed systems, as ground-truth transcripts are unavailable at runtime. This observation effectively nullifies the utility of the WER as a live diagnostic tool. By instead producing a product for transcription accuracy, our framework bridges this gap. Ultimately, this work provides the foundations for anytime-valid safety evaluations of high-dimensional discrete outputs, ith potential for extension to machine translation, code generation, and autonomous command-and-control ecosystems where reliability is paramount.

\FloatBarrier
\newpage

\section*{Acknowledgments}

This work was supported by the Australian Defence Science and Technology (DST) Group via the Advanced Strategic Capabilities Accelerator (ASCA) program.

\section*{Impact Statement}

This work explores the potential for enhancing robustness in adversarially exposed acoustic systems, specifically Automatic Speech Recognition models, and exists within the oeuvre of Adversarial Machine Learning. While defensive works in this space are typically assessed as eliciting no harms, we feel it is important to emphasize two key societal concerns which may be of note. 

The first of which is that there are some applications where a lack of robustness in a model may be positive. In a world where broad scale surveillance is increasingly normalized, it may well be the case that adversarial attacks may induce privacy, creating a net public good. 

The second relates to how works like this position risk and harm. A common precept within the Adversarial Machine Learning community is to assume a particular threat model, with the nature of academic comparisons often incentivizing us to then follow in the footsteps of those who came before us. However, in doing so, we inadvertently create---and, crucially, present----a myopic view of the risk landscape. In essence, we portray to practitioners that risk is concentrated within the areas in which we act as a community, when our investigations may be more motivated by historic alignment to academic norms and mathematical convenience. This work considers $\ell_2$ perturbations, which, while aligned with classical acoustic threat models, still represent a restriction relative to the overall threat landscape. Such a consideration is especially important, given the limitations associated with $\alpha_{atomic}$, as discussed within the work.

We emphasize the above points not just for the risks of erroneous portrayals of risk to practitioners, but also because our focus on these spaces inherently biases real attacker behavior away from these threat models. After all, if an attacker understands that an $\ell_2$ threat model is likely defended against, they're naturally incentivized to consider an alternative pathway for model manipulation. 

With these points made, we still believe that research into defences, and in particular certified defences, induce a net societal gain. Improving robustness to natural or adversarial perturbations will improve the performance of systems that are already one of the dominant access portals for AI within the community. Moreover, voice-based systems provide significant accessibility dividends to members of the community who do not have the ability to employ textual impacts, presenting an additional accessibility dividend. 

\bibliographystyle{icml2026}
\bibliography{references}

\appendix 

\section{Algorithm}\label{app:algorithm}

We present our full ASR certification pipeline in Algorithm~\ref{alg:pipeline}. In practice, we set $\tau = 0.5$, corresponding to majority occurrence under the smoothing distribution.

\begin{algorithm}
\caption{Certified ASR Pipeline}\label{alg:pipeline}
\begin{algorithmic}[1]
\State \textbf{Input:} Audio signal $x$, target SNR (dB), confidence level $\alpha = \alpha_{atomic} + \alpha_{tourn}$, base model $f$, threshold $\tau \in (0,1)$, step size $\lambda \in (0,1]$
\State \textbf{S0 (Noise Calibration):}
\State Compute signal power $P_s = \frac{1}{d} \sum_{i=1}^d x_i^2$, noise power $P_\epsilon = P_s / 10^{\text{SNR}/10}$, and set $\sigma = \sqrt{P_\epsilon}$.

\State \textbf{S1 (Candidate Vocabulary Construction):}
\For{$i = 1$ to $N_1$}
    \State Sample $\epsilon_i \sim \mathcal{N}(0, \sigma^2 I)$
    \State $Y_i \gets f(x + \epsilon_i)$ \hfill // $Y_i \in \mathcal{V}^\star$
\EndFor
\State $\mathcal{V} \gets \bigcup_{i=1}^{N_1} \{ w : w \in Y_i \}$

\State \textbf{S2 (Atomic Certification):}
\For{$t = 1$ to $N_C$}
    \State Sample $\epsilon_t \sim \mathcal{N}(0, \sigma^2 I)$
    \State $Y_t \gets f(x + \epsilon_t)$
    \For{each $w \in \mathcal{V}$}
        \State $Z_{w,t} \gets \mathbb{I}\{ w \in Y_t \}$
        \State $E_{pos}(w) \gets E_{pos}(w) \cdot (1 + \lambda(Z_{w,t} - \tau))$
        \State $E_{neg}(w) \gets E_{neg}(w) \cdot (1 + \lambda(\tau - Z_{w,t}))$
    \EndFor
\EndFor

\State $\mathcal{V}_{cert} \gets \{ w \in \mathcal{V} : E_{pos}(w) \ge 1/\alpha_{atomic} \}$
\State $\mathcal{V}_{excl} \gets \{ w \in \mathcal{V} : E_{neg}(w) \ge 1/\alpha_{atomic} \}$

\State \textbf{S3 (Filtered Transcription Sampling):}
\For{$i = 1$ to $N_3$}
    \State Sample $\epsilon_i \sim \mathcal{N}(0, \sigma^2 I)$
    \State $Y_i \gets f(x + \epsilon_i)$
    \State $\tilde{Y}_i \gets (w \in Y_i \;\text{s.t.}\; w \in \mathcal{V}_{cert})$
\EndFor
\State Select top-$K$ most frequent unique sequences from $\{\tilde{Y}_i\}$ as candidates $\mathcal{C}$

\State \textbf{S4 (Tournament Certification):}
\State Initialize wealth $E_i \gets 1$ for each $C_i \in \mathcal{C}$
\While{$\max_i E_i < 1/\alpha_{tourn}$ and $t < N_3$}
    \State Sample $\epsilon_t \sim \mathcal{N}(0, \sigma^2 I)$
    \State $Y_t \gets f(x + \epsilon_t)$
    \State $\tilde{Y}_t \gets (w \in Y_t \;\text{s.t.}\; w \in \mathcal{V}_{cert})$
    \State $i^* \gets \arg\min_i \text{WER}(C_i, \tilde{Y}_t)$
    \For{$i = 1$ to $K$}
        \State $E_i \gets E_i \cdot (1 + \lambda(\mathbb{I}(i = i^*) - 1/K))$
    \EndFor
\EndWhile

\State \textbf{Output:} $\hat{Y} = \arg\max_{i} E_{i}, \quad R = \min\{R_{atomic}, R_{tourn}\}$
\end{algorithmic}
\end{algorithm}

\subsection{Configuration}

Our evaluation was performed across four ASR architectures, to demonstrate the model-agnostic utility of our approach. These include Whisper (Large-v3 \& Small, MIT License, \citet{radford2023robust}), HuBERT-Large (MIT License, \citet{hsu2021hubert}) and Wav2Vec2-Large (MIT License, \citet{baevski2020wav2vec}). Of these, the Whisper is a transformer-based encoder-decoded model; HuBERT is a self-supervised hidden unit BERT model fine-tuned for CTC-based ASR; and Wav2Vec is a self-supervised framework. Evaluations are performed zero-shot (inference-only) using the \textit{test-clean} and \textit{test-other} subsets of the \textbf{LibriSpeech} dataset~\citep{panayotov2015librispeech} (licensed under Creative Commons Attribution 4.0 International) and the English \textit{test} split of the \textbf{Common Voice 17.0} variant~\citep{ardila2020common}, which employs the Mozilla Public License 2.0. For each permutation, we process 100 random sentences, resulting in a comprehensive matrix of 3,200 unique certification trials. 

All experiments were conducted using the parameters outlined in Table~\ref{tab:hyperparams}, on a H100 Tensor Core GPU with 80GB VRAM, using 16-bit precision for Whisper-Large-v3. Total experimental time was $8$ GPU-days. 

\begin{table}[h]
\centering
\caption{Hyperparameter Configuration for Tournament Framework.}
\label{tab:hyperparams}
\begin{tabular}{llc}
\toprule
Stage & Parameter & Value \\
\midrule
Discovery (S1) & Sample Count ($N_{S1}$) & 50 \\
Certification (S2) & Sample Count ($N_{S2}$) & 1000 \\
Nomination (S3) & Candidate Count ($K$) & 5 \\
Tournament (S4) & Max Sample Budget ($N_{S4}$) & 250 \\
\midrule
Martingale & Confidence Level ($\alpha$) & 0.01 \\
Betting & Instance Count Parameter ($\lambda_{inst}$) & 0.50 \\
Betting & Tournament Parameter ($\lambda_{tourney}$) & 0.20 \\
\bottomrule
\end{tabular}
\end{table}

\paragraph{Noise Augmentation} Our certification framework is built upon Additive White Gaussian Noise (AWGN) at four Signal-to-Noise Ratio (SNR) levels: $\{10, 5, 0, -5\}$ dB. Given a clean audio signal $x$, the noisy realization $x'$ is generated as $x' = x + \eta$, where $\eta \sim \mathcal{N}(0, \sigma^2)$ and $\sigma^2$ is scaled to achieve the target SNR. All audio is standardized to a single-channel 16kHz format before transcription.

\section{Linguistic Fragility}
\label{app:POS}

In a linguistic context, robustness of the final transcription may not be the only goal. In fact, in security conscious environments, it may also be important to audit the robustness of individual words. Such an audit framework also provides key information to those developing and deploying ASR models, as the results can be leveraged to understand weak points in systemic performance.

\begin{table}[h]
\centering
\caption{Definitions and examples for Part-of-Speech (POS) categories used in the linguistic fragility audit.}
\label{tab:pos_definitions}
\begin{small}
\begin{tabular}{lll}
\toprule
Tag & Category & Examples \\
\midrule
PRON & Pronoun & I, you, he, she, they, it \\
CCONJ & Coordinating Conjunction & and, or, but, so, yet \\
ADP & Adposition (Preposition) & in, on, at, by, with, from \\
AUX & Auxiliary Verb & is, are, was, were, has, have \\
ADV & Adverb & quickly, very, well, yesterday \\
ADJ & Adjective & big, blue, happy, loud \\
VERB & Verb & run, jump, think, speak \\
NOUN & Common Noun & dog, house, table, idea \\
PROPN & Proper Noun & London, Alice, Google, Monday \\
PART & Particle & 's, not, to (infinitive) \\
\bottomrule
\end{tabular}
\end{small}
\end{table}

To achieve this audit, we employed spaCy Natural Language Processing framework~\citep{honnibal2020spacy}, where each word was tagged to a Part-of-Speech (POS) category (see Table~\ref{tab:pos_definitions}) using spaCy's \texttt{en\_core\_web\_sm} model. We then calculated the \textit{Certification Recall}—defined as the percentage of tokens within a given linguistic class that successfully accumulated enough statistical wealth to be formally certified.

Our audit indicates that robustness is not uniformly distributed across word classes. Functional tokens, such as Pronouns (PRON, 21.7\% recall), Conjunctions (CCONJ, 16.7\%), and Adpositions (ADP, 13.1\%), are significantly easier to certify in noise. In stark contrast, content-heavy tokens, such as Nouns (NOUN, 2.8\% recall), Verbs (VERB, 3.8\%), and Proper Nouns (PROPN, 1.2\%), exhibit extreme fragility, with recall rates falling drastically.

This discrepancy likely stems from the relative semantic density of these content words in the training corpora for these models. In noisy environments, an ASR model may hallucinate numerous phonetically similar but semantically distinct nouns (e.g., "cat", "cap", "bat"). This high entropy disperses the probability mass, preventing any single content token from accumulating the statistical wealth required for certification under a strict E-value martingale. Functional words, however, belong to smaller, more closed linguistic classes and are often highly predictable given the surrounding context, allowing them to rapidly gain consensus. This audit highlights the need for "linguistically-aware" certification frameworks that can prioritize or differentially weight evidence accumulation for critical content words to improve system-level trust.

\begin{figure}[t]
    \centering
    \includegraphics[width=\columnwidth]{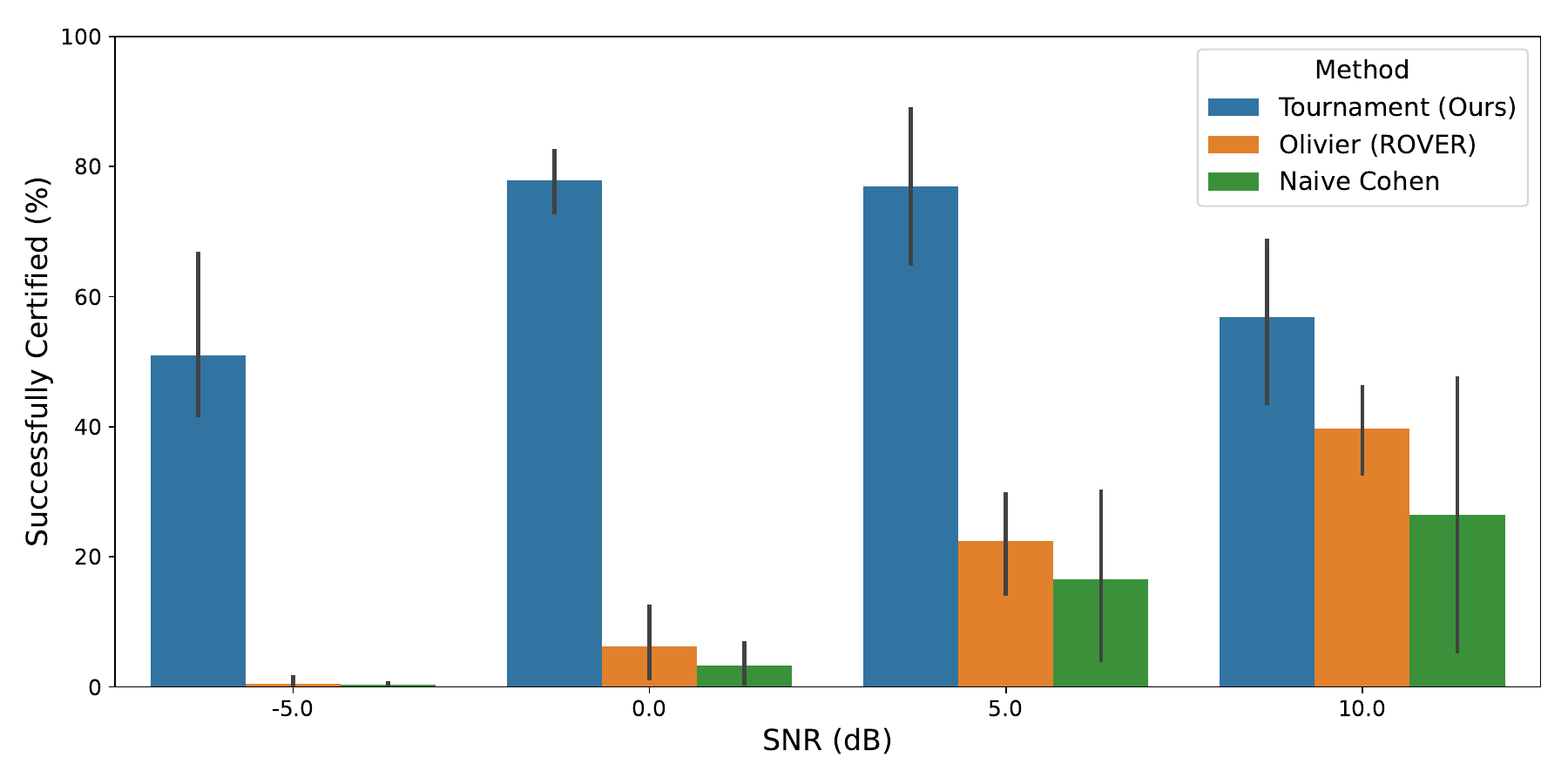}
    \caption{Relative Certification performance across all approaches}
    \label{fig:certification_deltas}
\end{figure}

\section{Additional Results}\label{app:further_results}

\begin{figure}[t]
    \centering
    \includegraphics[width=0.8\columnwidth]{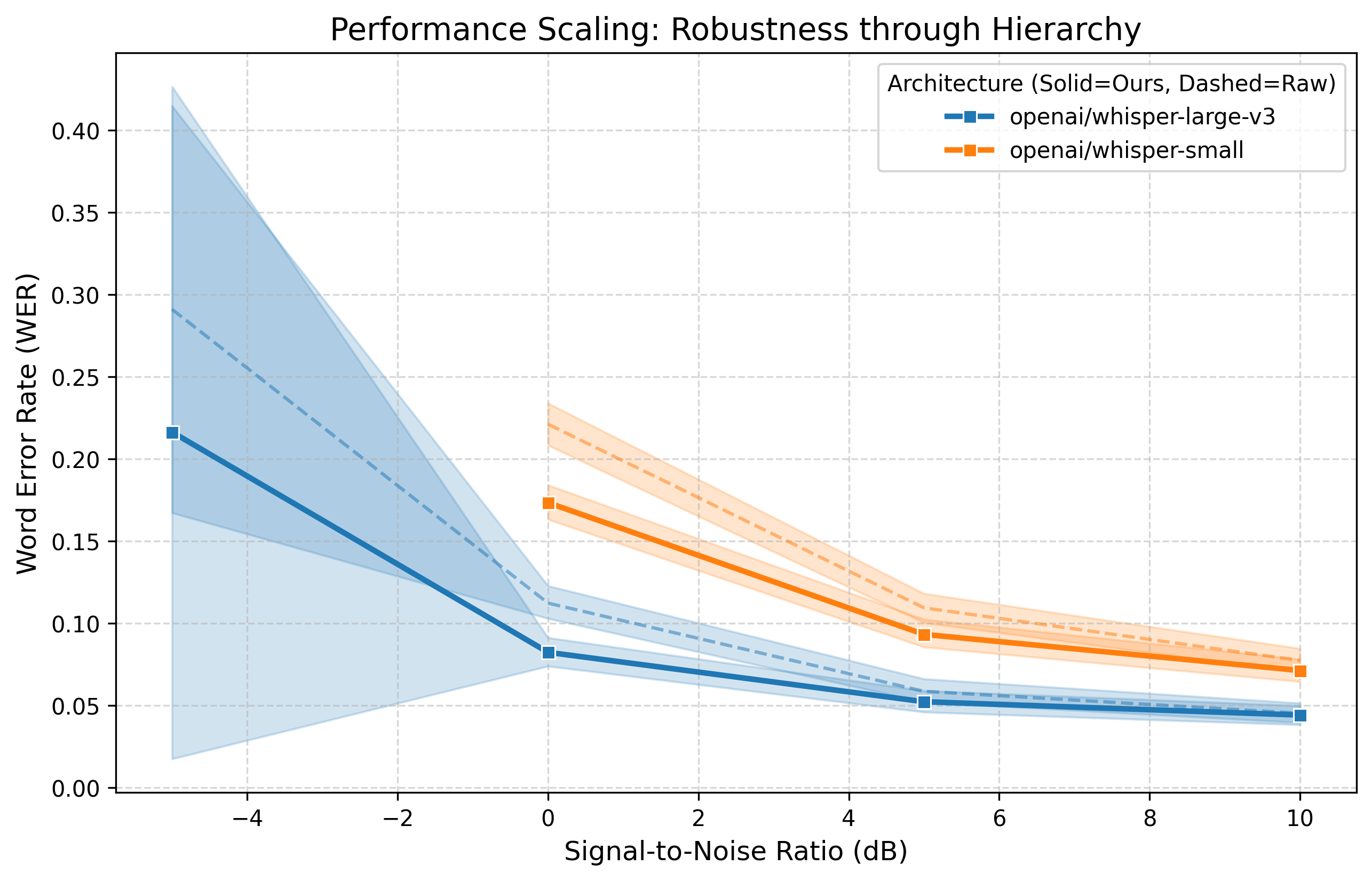}
    \caption{Relationship between SNR and WER. Solid lines: Certified Transcriptions, Dashed lines: Uncertified Transcriptions Under Noise}
    \label{fig:performance}
\end{figure}

\begin{figure}[t]
    \centering
    \includegraphics[width=\columnwidth]{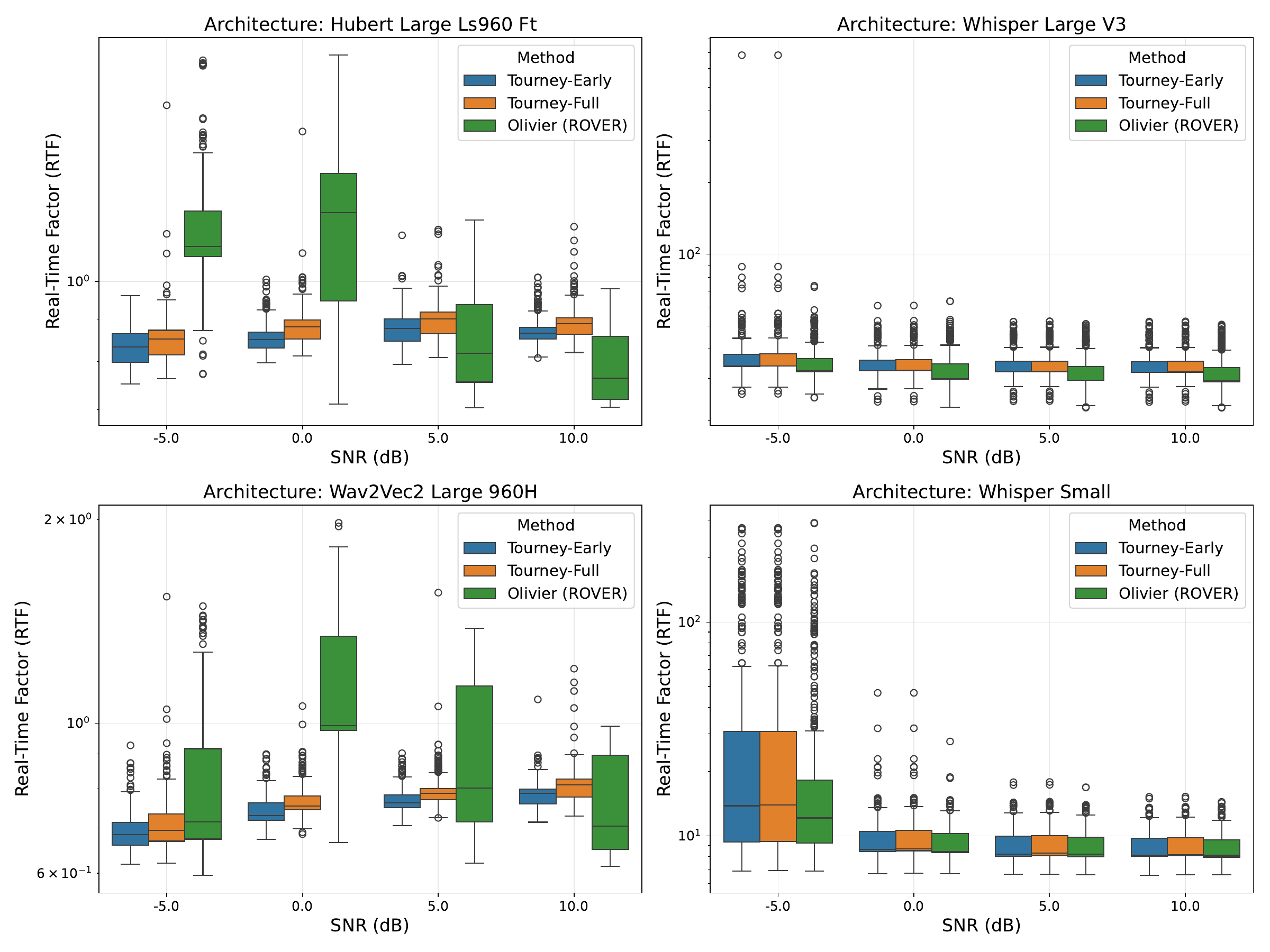}
    \caption{Relationship between SNR and the Real Time Factor (RTF) for different models.}
    \label{fig:rtf}
\end{figure}

The performance deltas visualized in Figures~\ref{fig:performance} and \ref{fig:certification_deltas} confirm that the utility of the Tournament framework extends beyond mere certification. By aggregating evidence across a massive search space, the system consistently produces a certified transcription that is significantly more accurate than any individual transcription under noise. As detailed in Table~\ref{tab:final_master}, this multiplicity subsidy is most pronounced in high-noise environments, where our framework achieves an absolute WER reduction of up to 10.6\% at SNR -5dB on the LibriSpeech dataset, and 8.1\% on Common Voice. This indicates that the aggregation mechanism is particularly valuable precisely when the underlying model begins to fail catastrophically. We note that the relative performance differences seen between the CTC-based architectures (HuBERT, wav2vec) and the auto-regressive models of Whisper are not distributed evenly (see Figure~\ref{fig:rtf}). We believe that this is a product of both the relative balance of the models computational cost to the certifications, and, perhaps more importantly, a decreased sensitivity to noise, which manifests as significantly smaller transcription sets for the ROVER based mechanism to aggregate ov

A critical requirement for practical ASR safety is real-time or near real-time feasibility. Table~\ref{tab:decision_time} demonstrates that for CTC-based architectures (HuBERT, Wav2Vec2), our anytime-valid framework is \textbf{17--20\% faster} than traditional ROVER alignment. While auto-regressive models like Whisper-Large incur a higher absolute cost due to their decoding mechanisms, the anytime-stopping rule provides a \textbf{28\% compute saving} relative to a fixed-budget approach, exiting the tournament as soon as statistical confidence is reached. This proves that martingale-based certification is a practical path for sequence-level guarantees.

\begin{table}[h]
\centering
\caption{Computational Efficiency (RTF). Tournament anytime-stopping maintain competitive scaling even under heavy noise.}
\label{tab:decision_time}
\begin{small}
\begin{tabular}{l ccc}
\toprule
Architecture & Naive Cohen & ROVER & \textbf{Ours (EE)} \\ 
\midrule
HuBERT-large & 0.05x & 0.07x & \textbf{0.06x} \\
wav2vec2-large & 0.05x & 0.06x & \textbf{0.05x} \\
Whisper-large-v3 & 2.20x & 2.12x & \textbf{2.37x} \\
Whisper-small & 0.90x & 0.83x & \textbf{0.96x} \\
\bottomrule
\end{tabular}
\end{small}
\end{table}

\FloatBarrier

\end{document}